\documentclass{article} 
\usepackage{iclr2021_conference,times}

\usepackage{amsmath,amsfonts,bm}

\def\eqref#1{equation~\ref{#1}}

\def\1{\bm{1}}

\DeclareMathAlphabet{\mathsfit}{\encodingdefault}{\sfdefault}{m}{sl}
\SetMathAlphabet{\mathsfit}{bold}{\encodingdefault}{\sfdefault}{bx}{n}

\usepackage{hyperref}
\usepackage{url}

\title{Metaphors we learn by}

\author{Roland Memisevic\\
Qualcomm AI Research, Canada\thanks{Qualcomm AI Research is an initiative of Qualcomm Technologies, Inc.}\\
\texttt{rmemisev@qti.qualcomm.com} \\
}

\usepackage{comment}

\begin{document}

\maketitle

\begin{abstract}
Gradient based learning using error back-propagation (``backprop'') is a well-known contributor to much of the recent progress in AI. A less obvious, but arguably equally important, ingredient is parameter sharing -- most well-known in the context of convolutional networks. In this essay we relate parameter sharing (``weight sharing'') to analogy making and the school of thought of cognitive metaphor. We discuss how recurrent and auto-regressive models can be thought of as extending analogy making from static features to dynamic skills and procedures. We also discuss corollaries of this perspective, for example, how it can challenge the currently entrenched dichotomy between connectionist and ``classic'' rule-based views of computation. 
\end{abstract}

\section{Parameter sharing in AI} 
It is well-known that neural networks, regardless whether training is supervised or self-supervised, 
require large amounts of training data to work well. 
In fact, the ability to generalize requires the ratio 
\begin{equation}
\label{numparamsovernumexamples}
\frac{\# \mathrm{training\ examples}}{\# \mathrm{tunable\ parameters}}
\end{equation}
to be large \citep{hastie01statisticallearning}. 
To ensure generalization,  
one can maximize the number of training examples, 
minimize the number of tunable parameters, or do both. 
Parameter sharing is a common principle to reduce the number of tunable parameters 
without having to reduce the number of actual parameters (synaptic connections) in the network. 
In fact, it is hard to find any neural network architecture in the literature, 
that does not make use of parameter sharing in some way. 
The core theme of this essay is that parameter sharing, and its fundamental role in 
AI, is an instance of a long-held view of cognition in the school of thought 
of conceptual metaphor: that \emph{cognition is analogy making}  \citep{hofstadter2013surfaces,lakoff1980metaphors}. 
As AI capabilities evolve from low-level perception towards 
higher levels of cognition, cognitive metaphor may therefore play 
an increasingly important role in developing and training of AI models.

The most well-known example of parameter sharing is convolution. 
Convolutional networks re-use local receptive fields to exploit translation 
invariance in the data. This has an enormous effect on parameter count: 
by leveraging locality as an inductive bias, convolution allows us to reduce the 
number of network parameters by several orders of magnitude  
in just the first layer of a neural network applied to 
even a moderate image size\footnote{Consider the following 
back-of-the-envelope calculation: A typical image-crop used for training on images from 
Imagenet \citep{imagenet_cvpr09} is a $224\times 224$ pixels RGB image, 
leading to $224\times224\times3=150528$ inputs to the network. 
We get a lower bound on a "reasonable" number of hidden units in the first layer of a neural network by using as many hidden units in the first layer as number of pixels. 
We would prefer an over-complete basis, so this is a lower bound only. 
The number of parameters in that layer (not counting biases) would thus be $\approx150528^2\approx23$ billion. 
Convolution amounts to using local receptive fields and weight sharing. 
Using local receptive fields amounts to connecting each hidden unit to a 
locally confined region in the image. Although most current convolutional networks 
use small receptive fields, we can get an upper bound on the number of parameters 
in the convolutional layer by assuming large receptive fields of, say, size $10\times10$ 
which, accounting for RGB channels, would amounts to $300$ parameters per hidden unit. 
This would naively reduce the number of parameters in the network from $23$ billion to $150528\times300\approx4.5$ million, or approximately three orders of magnitude less. 
However, to account for the fact that we would expect similar $10\times10$ features in different regions in the 
image, we would probably want to use a significantly overcomplete set of such local features to cover every 
region in the image. While done naively, this would drive the number of parameters back up, 
weight sharing (using the same set of parameters in different regions of the image) allows us to 
have the cake and eat it to: 
If we use, say, as many hidden units as there are pixels within a receptive field, 
and use weight-sharing to apply the same receptive field everywhere in the image, we end up 
with $224\times224=50176$ hidden units (assuming padding) but only 
$300\times300=90000$ parameters in total. This is a heavily overcomplete basis, yet an 
even a further reduction in parameter count by another $1.5$ orders of magnitude.}. 
While convolutional is the most well-know example of weight sharing, there are many others. 
Sharing weights through time in a recurrent network is another well-known 
example, which is closely related to convolution (in fact, many recurrent networks are 
equivalent to a 1-d convolutional networks operating in time). 
A recurrent network can also be thought of as multi-layer networks with weights shared 
across timesteps. Since recurrent networks are typically trained via unrolling and 
back-propagation-through-time, it is customary to think of them as 
multi-layer networks with weight sharing between the layers.

\subsection{Deep learning as parameter sharing}
\label{deeplearningasparametersharing}
We can also view the practice of layering networks, even in the 
absence of any weight sharing between layers, as an instance of weight sharing: 
In a simple multi-layer network, any neuron in layer $L$ will share all 
the parameters up to layer $L-1$ with all the other neurons in layers $L$ 
and above. 
The same is true of the activations, which are the result of the computations that 
the weights subserve. 
More generally, a distinguishing feature (perhaps \emph{the} distinguishing 
feature) of any hierarchical structure, not just a neural network, is that the 
higher-level entities in the hierarchy share among each other the information 
contained in the lower-level entities. 
While sharing can simplify engineering in general hierarchical structures, the distinguishing 
feature of neural networks is that here it simplifies learning. 

Sharing in a hiercharchical structure goes hand-in-hand with compositionality: 
By definition, ``sharing'' requires at least two entities which do ``share''. 
In a hiercharchy that is $3$ or more layers deep, any interior layers are 
therefore necessarily compositional or, in the case of a neural network, constitute a 
distributed representation. A corollary is that any information processing in the interior is many-to-many. 
Any distributed (``factorial'') representation inside the neural network constitutes information sharing 
as well, since all states of one variable are shared between all states of another variable. 
In other words, a benefit of compositionality itself is that it facilitates information sharing.

In the absence of any external information (including any internal noise), 
no information is created when traversing a hierarchy, 
so hierarchies typically get "thinner" towards the top. 
Pooling layers in a convolutional network accompany the information loss by 
an actual reduction in the number of neurons. 
In many other cases, the information loss is accompanied by increasingly sparse 
representations as we move up the network. 
The last layer of a classifier network is in a sense the logical culmination, 
where one-hot vectors reduce the remaining information content to the bare minimum 
that is still considered useful for the task at hand. 
On the other hand, the amount of information that went in training the parameters 
determining a neuron's activation \emph{grows} as we move up the hierarchy. 
In that sense, the level of abstraction at which a neuron encodes information 
is correlated to (if not equivalent to) the amount of training data it has been 
exposed to. 

Neurons higher up in the network encode representations that are also 
increasingly \emph{disentangled} (see, for example, \cite{bengio2017consciousness}). 
For example, thanks to sparse distributed representations 
in higher layers of a network, a ``read-out'' layer can use a simple computation 
(the last layer in a neural network classifier being a simple linear model is the 
most well-known example). 
Viewed through the lens of weight sharing, such disentangling is not necessary per se or in any deep sense. 
The simplicity of the read-out follows from the fact that there is no coupling necessary between the 
output-units (logits). The units do not need to ``know'' about each other, so 
they can perform computations independently of one another. 
This, however, one could argue is true simply because they \emph{share} the underlying 
input vector (the pen-ultimate layer in the network), and along with it any computations 
contributing to it, which may include any required non-linear couplings. 

\subsection{Transfer learning as parameter sharing}
To improve generalization, instead of reducing the number of parameters (the denominator) 
in Equation\ref{numparamsovernumexamples}, we can increase the number of training examples. 
This proposition is trivial when considering a fixed network trained for a fixed task. 
However, many superficially unrelated tasks can be deeply ``structurally'' related. 
For example, it is well-known that across a vast number of computer vision tasks, supervised 
training  leads to Gabor-like features in the lowest layer of a convolutional network, rendering 
these ``universal'' in a sense. 
Higher-level features, similarly, can learn concepts that apply across a wide range of tasks 
(see, for example, \cite{girshick2014rich}). 

Transfer learning, that is, sharing a subset of a network's layers between tasks, 
exploits similarities between tasks, which amounts to effectively increasing the amount 
of training data. 
Perhaps the simplest form of transfer learning, which became popular in computer vision 
around 2014 (e.g., \cite{razavian2014}, \cite{donahue2014decaf}), amounts to 
pre-training a network on a large dataset (like ImageNet \citep{imagenet_cvpr09})
and then fine-tuning a subset of the parameters (in many case the last layer) 
on a data-scarce target task. 
This can also be viewed as using the pre-trained neural network as feature extractor for 
a (linear) model trained on the target task.  
This approach to transfer learning has since become popular in many tasks. 
And architectures containing a single, shared backbone network, feeding into multiple, task-specific prediction heads are now emerging as a standard approach in many applications.

After ImageNet-pretrained networks started to get applied in a variety of target tasks 
via transfer learning, which started around 2013, 
a common assumption has been that the relatively large number of classes ($1000$) 
plays a crucial role in the ability to learn generic, and thus transferable, 
features\footnote{
One could view self-supervised learning as an extreme-case of transfer learning where the  
source task comes with a very wide variety of labels. 
In particular, reconstruction-based models like autoencoders can be thought of as learning 
from a combinatorial 
number of possible outputs (the reconstructed input). 
Self-supervised learning relies crucially on confining the training data to be drawn from a small 
subset of the 
data domain a model is trained on, such as human-generated text for language models or natural images for vision models.
In fact, it is widely agreed that training a vision-model on images with random iid pixels would not be useful. 
Even synthetic random images need to exhibit structure to constitute useful training signals for downstream tasks as discussed in \citet{lookingatnoise}. 
One motivation for using self-supervised learning is that to 
generate perfect natural images, 
a model will need to know everything about the physical processes that gives rise to the 
images, hence be a perfect model of the world. 
It is like using a source task that is so broad that it will encompass any target task of interest. 
While this is true in principle, it comes with the downside that model capacity and 
source training set size may need to be extraordinarily large. 
In contrast to unsupervised pre-training, transfer learning makes the selection 
of an appropriate set of source tasks to solve a given target task a key component of 
the model development effort. 
}. 
\citet{armand_weakly}, for example, showed how image captions can be an effective training data source for learning generic features. 
A quantitative study that aimed at confirming the dependence of transfer learning
performance on source task granularity is our work in \citet{mahdisoltanietal2018}.

\subsubsection{Data-centric learning versus data-centric pre-training}
The well-known need for training data 
(to minimize the denominator in Eq. \ref{numparamsovernumexamples}) 
is also reflected in recent movements towards data-centric learning 
(for example, \citet{ng2021mlops}). 
An important nuance is that there are two fairly distinct  
ways to generate and manage the data needed for learning. 
The first, collecting training data for any particular task at 
hand, and then training a network to solve that task, 
faces the issue that the learned capabilities tend to be 
highly vertical and narrow. 
This means that an elaborate data pipeline needs to be built 
\emph{for every application} requiring substantial amounts of 
engineering and development. 
The problem is aggravated by the fact that AI use-cases (and 
arguably cognitive capabilities more broadly)  
are ``tail-events'': there is a myriad of possible applications, 
each one coming with its own peculiarities and requirements. 
This problem is related to the out-of-distribution (OOD) problem in machine learning, which refers to the
fact that training a model to show any reasonable behavior on training data drawn outside the support 
of the distribution it has been trained on is hard.

The second way in which we can bring to bear data-centric learning 
is to target a generalist, multi-purpose AI model, that can 
develop meaningful representations across a large set of potential capabilities 
and use-cases. This solution also relies on data generation as the driving force. 
However, data generation is not viewed as task-specific requirement, 
but as a way to instill a broad set of capabilities within a broad application domain. 
This way of performing data-centric machine learning is very different from 
the application-specific approach discussed above in it treats any desired capabilities 
as \emph{emergent phenomena}. 
An example is the use of an ImageNet-pretrained network as a generic visual feature 
extractor mentioned above. 
Another, more recent, example is our work on the ``something-something''
task \citep{somethingsomething}, where the goal is to generate textual descriptions of 
events involving objects in videos. The purpose of that task is \emph{not} to solve any 
particular use-case, but to let broad low-level visual capabilities, 
such as the detection and tracking of objectse emerge in response to training. 
A third, even more recent, example is the training of auto-regressive language models 
on text for the emergence of generic language processing capabilities 
\citep{radford2019language}. 

\section{Analogy making as parameter sharing} 
The representation of one concept in terms of another, related concept 
is known as \emph{metaphor}. Traditionally, metaphors have been regarded as 
figures of speech in linguistics and literature. Accordingly they are 
often viewed as creative devices related to an artistic use of language. 

However, more recently, specifically in the area of cognitive linguistics, 
metaphors have been argued to play a key role in human-like 
intelligence and thought (e.g., \citet{lakoff1980metaphors}). 
This has given rise to the school of thought of \emph{conceptual metaphor} 
\citep{wikipedia:conceptualmetaphor} which argues that metaphors are 
deeply pervasive in, and structure every aspect of, human cognition. 
According to this view, the meaning of any concept or linguistic expression 
is rooted in its relationship with other concepts. 

For example, the fact that the word ``argument'' is conceptualized in terms 
of ``war'' in some languages has the effect that concepts 
of ``winning'', ``attacking'' and ``strategy'' can structure our thoughts about arguments. 
Or the fact that ``time'' can be equated to ``money'' in some languages, 
lets concepts of ``wasting'', ``spending'', and ``saving'' structure our use of 
the word in those languages. 
\citet{lakoff1980metaphors} argue that examples like these are not rare 
examples of the creative use of metaphors, but a reflection of the core 
metaphoric nature of high-level cognition itself. 

Douglas Hofstadter, in numerous writings since the 1980's, 
has been going further, by arguing that analogy making (which we 
shall use interchangeably with metaphor) 
is the driving force of not only abstract, high-level cognition, 
but the key to essentially all of human intelligence (see, for example, \citet{seeingasandseeingas,hofstadter1995fluid,hofstadter2013surfaces}). 
According to this view, capabilities ranging from simple object 
categorization to creating novel analogies in literature 
and science, are expressions of one single underlying 
principle -- that of analogy making. 
Accordingly, in this view, metaphors are created ``numerous times every a second'' \citep{hofstadter2013surfaces}. 
A corollary is that metaphors are not confined to being verbal 
devices. Rather, human-like cognition is based upon both verbal and 
non-verbal (or ``pre-verbal'') metaphors \citep{hofstadter2013surfaces}.  

The pervasiveness of parameter sharing in machine learning 
suggests interpreting \mbox{Hofstadter's} all-encompassing view 
of analogy making as a way to reuse neural circuitry --  
in other words as weight sharing. 
From this perspective, analogy making is a necessity to enable learning and 
generalization. Conversely, the importance of weight sharing  
may also explain why the use of higher-level metaphors is so pervasive 
in cognition and thinking.

\subsection{Recurrent networks and analogies between skills}
Metaphor as a mechanism to lend meaning to concepts not only applies to 
``nouns'', or objects, but equally to ``verbs'', activities, procedures and 
more broadly to any kind of experience in the widest sense \citep{hofstadter2013surfaces}. 
In the following we shall focus on a particular feature to classify 
the use of metaphors by: the distinction between \emph{static} and \emph{sequential}. 

We first note that any sequential form of information processing, like  
that inherent in classic (Von-Neumann) models of compute, implies a form 
of information sharing. 
Specifically, in the same way that neurons in one layer of a neural network share 
the results of computations performed by the layers below, 
computations in a classic computer \emph{share} results from earlier computations. 
In that sense, we could view sharing of computations as a key ``design principle'' 
and perhaps motivation for classic compute models to be sequential in the first place. 
This kind of sharing in classic compute does not facilitate learning. 
But it facilitates the reuse of hand-crafted computational mechanisms, 
such as routines, functions, libraries, modules, packages, etc. 

In a recurrent network\footnote{More precisely, any network using weight-sharing across time.
We shall use ``recurrent network'' as an umbrella term to encompass any such network. This 
includes, in particular, auto-regressive models that generate one output at a time as a function of previously generated outputs.} 
(which is in a sense the connectionist ``counterpart'' of 
a classic computer), the same kind of sequential information sharing is present. 
But on top of facilitating the reuse of computations and compositionality, 
sharing here also facilitates generalization -- in other words, the acquisition 
through learning of the mechanisms underlying those computations. 

Recurrent (and auto-regressive) networks are much better 
suited to support transfer learning 
than feed-forward networks, because they allow for much more flexibility 
in the definition of any task. Recurrent networks are able to generate sequences. 
This renders them structured prediction 
models \citep{lecunbottoubengionhaffner,Lafferty2001ConditionalRF}, which 
can generate outputs from a combinatorial set of candidate outputs. 

Moreover, since recurrent networks emit outputs incrementally, the outputs can 
also be interpreted as actions performed in an environment. 
This allows a recurrent network to operate as an \emph{agent} interacting with 
an environment. 
This dramatically increases the number of tasks -- and supervision signals -- 
that can drive learning in a recurrent network. 
And the analogical representations that can thereby emerge through learning can include 
not only static features but also 
(dynamic) skills, procedures, approaches, solution strategies, and so on. 
From the perspective of analogy making, a neural network that operates sequentially 
can learn to represent a task or a sub-task \emph{``as''} a kind of task it is 
familiar with \citep{seeingasandseeingas}. The same is true of the computational 
or ``cognitive'' machinery it can use to solve it. 

In context of the OOD problem mentioned above, seeing a task ``as'' another, familiar 
task is like projecting the new task onto a ``subspace'' of tasks the network is able to solve. 
This can be a far reaching capability, as it can involve solving even tasks that are 
difficult to learn (for example, due to being non-differentiable) by interpeting them as 
other, sufficiently similar tasks, that are learnable (for example, thanks to being differentiable)
and that have thus been learned previously. 

Understanding the implications and effects of domain transfer between skills 
(as opposed to static features) is not very deeply explored and an open research endeavor.
A notable exception is the rapid recent progress on language-based reasoning tasks 
(commonly referred to as ``System-2'' tasks \citep{kahneman2011thinking}) 
via auto-regressive models. 
This includes the work by 
\citet{Recchia2021TeachingAL,lu2021pretrained,cobbe2021training,minerva2022,googlepalm,googlelamda}, 
that we shall elaborate on in the next section.

\section{Analogy making and the foundations of compute} 
Metaphors impose constraints on the understanding and the use of language, 
because they let a concept evoke other concepts. 
Hofstadter refers to the set of other concepts that any given concept can evoke 
as the concept's ``\emph{halo}'' \citep{seeingasandseeingas}. 
We can conceive of the halo as encompassing not just semantic information 
but every aspect that is activated by a given expression (or ``thought''), 
including, say, the tactile sensation of a physical object it involves, 
the sound of its pronunciation, the visual appearance of its spelling in a given font, etc. 
As such, the halo of a concept is multi-dimensional\footnote{
The notion of a concept's halo can add an important nuance to the question about  
the true level of understanding a neural network may exhibit: 
whether we consider a neural network to truly understand any given concept 
or not hinges on which aspects of that concept`s halo we 
define ``true understanding'' to include. 
A common argument against a neural network's ``true understanding'', for example, rests
on their lack of grounding. If the definition of ``true understanding'' is meant to include 
visual  representations, then most current language models are clearly 
far from truly understanding many concepts. And while this is beginning to change 
thanks to multi-modal models (for example, \citet{flamingo,gato}), there is still a 
long way to go. 
If we conceive  of ``true understanding'' to include just those aspects of 
a concept that facilitate some level of logical reasoning, 
then existing language models could very well be argued to truly understand 
many concepts and situations. 
For example, in the System-2 tasks discussed above, 
information derived from mere text, without any multi-modal grounding, 
is sufficient to solve a variety of mathematical challenges 
and word problems (for example, \citet{cobbe2021training,minerva2022}). 
Importantly, this line of work shows that language is able to force a neural network 
onto chains-of-thought that are not only well-formed syntactically, but that are 
also sensible semantically (despite being deprived of those aspects of semantics that 
are linked to sensory perception).}. 

In the following, we shall turn our attention to \emph{dead metaphors}, 
as discussed, for example, by \citet{travers1996programming}), which can be 
thought of as one dimension in the halo of a concept along which the concept exerts its 
influence.

Dead metaphors are metaphors that are so commonly used that they no 
longer ``feel'' metaphoric. 
In other words on the surface they seem not to evoke any reference concepts. 
An example of a dead metaphor is the expression ``time is running out': 
while it likely originated as a reference to the sand in an hourglass,  
the expression usually does not invokes that image today. This is what 
renders it a dead metaphor. 
Other examples of dead metaphors 
include the expressions ``hanging up the phone'', ``the legs of the table'' or 
``patching source code''. 

Dead metaphors allow us to make references to concepts implicitly and 
even unknowingly. 
However, it is important to note that they are based on more than mere memorization: 
while some of the aspects they refer to are lost, others persist. 
For example, while the 
expression ``time is running out'' may no longer invoke the image of an hour glass, 
it may still invoke concepts of \emph{depletion}, of \emph{physical inevitability}, 
and of an overall \emph{short time frame}. 
And other uses of the expression ''running out'', 
such as ''we are running put of pencils'', may invoke a 
subset of the same concepts (such as the concept of \emph{depletion}) while ignoring 
others (such as the concept of \emph{short time frame}). 

\subsection{Dead metaphors and classic compute}
The use of natural language could, to a first approximation, be  
said to subserve two fairly distinct purposes: 
communication and ``thinking''. 
The same can be said regarding the use of formal languages in 
classic compute. A programming language allows us to communicate 
instruction-sequences to the machine, and the same (or a 
lower-level, compiled) language allows the machine to execute 
(``think through'') these. 
In contrast to natural language, ``thinking'' in the machine 
leverages a highly formalized use of language that relies 
foremost on dead metaphors, specifically those based primarily 
on syntax. 
This includes classic programming concepts, 
like ``if/then'' to perform conditionals, 
``call/return'' to utilize aggregate functionality, 
or ``while''-constructs to evoke the repetitive execution of 
the same.

Training auto-regressive models on System-2 reasoning tasks 
also instills a degree of formal ``thinking'' in these models. 
Similarly, the computations performed in the models are governed 
by language, that structures the chains-of-thought leading 
to the correct solutions.
But the involved metaphors, while being less rich and creative than 
those used in open ended language generation, are at the same time 
significantly less rigid and syntactical than those used 
in classic compute. 
In other words, while they may be ``dead'', they are significantly less so than the metaphors 
underlying classic programming concepts. 

Consider, for example, the following math word problem from 
the GSM8K dataset \citep{cobbe2021training}, which auto-regressive 
models are now able to solve fairly well (e.g., \citet{wei2022chain}):

\begin{verbatim}
Jean has 30 lollipops. Jean eats 2 of the lollipops. With the 
remaining lollipops, Jean wants to package 2 lollipops in one 
bag. How many bags can Jean fill?
\end{verbatim}

To solve this problem, a model needs to understand that 
a remainder needs to be computed, and that the result needs to be divided 
by $2$. Distilling this kind of requirement from the question (and then 
performing the resulting computations) is obviously a more formal and 
straightforward process than everyday human thoughts. 
But it is also a significantly \emph{less} 
formal process than the execution of a piece of code (a task, 
which neural networks, however, can be trained to perform as 
well (see, for example, \citet{zaremba2014learning,nye2021show})).

This suggests rethinking the common dichotomy of computation 
as classic (or serial, symbolic, deliberate) on the one hand versus 
connectionist (or parallel, associative, intuitive) on the other. 
A more appropriate view of computation in light of neural networks' ability 
to execute System-2 thought processes may be that of a continuum, which  
ranges from \emph{classic-like} (mostly based on dead, formalistic metaphors) 
to \emph{human-like} (including rich and possibly creative metaphors). 
This novel, ``third'', compute paradigm encompasses both classic compute and deep learning. 

The computations in this paradigm are powered primarily by neural networks, running on 
parallel hardware. However, given a neural network's ability to execute classic computations as well, 
the parallel hardware can in principle support computations anywhere along that spectrum.

\subsection{Classic compute at a neural network's disposal}
There is another way in which transfer learning in auto-regressive (or recurrent) 
models can challenge our view  of the classic--connectionist dichotomy. 
Auto-regressive models, as discussed, emit outputs that can be interpreted as actions a 
model can take in an environment. 

Allowing a neural network to take actions in the environment 
is typically accomplished practically by letting 
the model use pre-defined language to communicate with its 
environment. In current implementations, the latter is a (``classic'') program, 
such as a script, that calls upon the auto-regressive model to produce outputs 
token-by-token. 

The information that the network can learn to communicate to 
its environment can include instructions to classic software 
running in the same environment. 
The environment can respond by writing information 
back into the language-buffer before continuing to step the model. 
This makes it possible to train models to control and use 
classic software, including calculators \citep{cobbe2021training}, 
web-browsers \citep{webgpt}, addressable external memory \citep{Recchia2021TeachingAL} or RL-environments \citep{decisiontransformer2021}.
\citet{Recchia2021TeachingAL} proposes the term  
\emph{environment forcing} to refer to this approach.

Environment forcing is a further challenge to the entrenched dichotomy of computation. 
We are used to thinking of neural networks as computational ``slaves'' 
that run on parallel accelerator hardware, 
which in turn is controlled by classic hardware. 
Due to the use of environment forcing, neural networks are increasingly trained to 
operate classic hardware instead -- 
letting the latter play the role of the computational slave. 
The use of a calculator in the context of solving math word problems \citep{cobbe2021training} 
is an example of this. 

Although environment forcing has thus far been mostly 
used to let neural networks operate ``simple'' software, 
such as web-browsers or  calculators, it is conceivable that 
the complexity of neural network-controlled software will grow significantly in the future. 
Such a trend could be fueled further by the emerging ability of 
neural networks to write code \citep{codex,alphacode}, allowing for 
situations where a network learns to write code to be executed 
on classic hardware, to then use the output for further processing. 
The use of a calculator in the context word problems \citep{cobbe2021training} is, 
in fact, a very simple example of this.

Environment forcing also enables a neural network to access long-range, 
persistent memory in a very different way than previous approaches, 
such as LSTM \citep{lstm}, 
Neural Turing Machines \citep{ntm2014}, 
or Memory Networks \citep{memorynetworks2014}.
These methods can be thought of as improving long-range memory  
by alleviating difficulties associated with back-propagating through many time-steps. 
In contrast to these methods, environment forcing makes it possible to use  
long-range memory without resorting to back-propagation-through-time altogether: 
some tokens (or token sequences) are simply reserved to denote memory read or write operations. 
Upon a model emitting any of these tokens, the code that calls the model's inference method 
performs the appropriate operations on an extenral (classic) 
memory (see, for example, \citet{Recchia2021TeachingAL}. 
For read-operations (or any other operations that should influence the model's internal ``state''), 
the code writes resulting values back into the model`s token-stream. 
In that way, environment forcing separates the task of 
memory access into two distinct sub-tasks: 
(i) persistent information storage and 
(ii) a policy deciding when and how to write to or read from memory. 
This is different from the previous approaches such as LSTM, 
which amount to training the neural network to solve both (thereby incurring 
dependence on back-propagation to achieve storage persistence). 
Environment forcing instead leverages the neural network to learn the memory 
access policy but relegates the storage itself to classic compute hardware.

At first sight, 
combining neural language models with classic compute, 
for the purpose of memory or computation, 
is reminiscent of hybrid neuro-symbolic approaches. 
It is important to note, however, that environment forcing does not amount to performing any AI 
related tasks in the classic hardware. The classic hardware is merely performing what classic hardware 
is good at: storage, simple calculations, etc. Any sophisticated inference, and in particular, the 
control policy with which the classic hardware is operated, is left 
to the neural network to learn. 

\section{Discussion}
As the scope of emergent skills in neural networks grows, 
the theory of analogy making and conceptual metaphor may become increasingly relevant, 
since training requires attention to the choice of pre-training tasks that 
may instill any desired capability in a model. 
This turns machine learning from an area dominated by neural architecture 
development into more of an ``educational'' practice, that relies on a deep understanding 
of the relationships between concepts, skills and tasks. 
In that way, conceptual metaphor and the study of analogy making have not 
only foreshadowed developments in AI but can also play a key role in 
driving its development going forward.

\bibliography{sharing}
\bibliographystyle{sharing}

\appendix

\end{document}